\documentclass[letterpaper, 10 pt, conference]{ieeeconf}  

\IEEEoverridecommandlockouts                              

\usepackage{cite}
\usepackage{amsmath,amssymb,amsfonts}
\usepackage{algorithmic}
\usepackage{graphicx}
\usepackage{textcomp}
\usepackage{xcolor}
\usepackage{array}
\usepackage{makecell}
\usepackage{multirow}
\usepackage{threeparttable}
\usepackage{wrapfig}
\usepackage{hyperref}
\usepackage{subcaption}
\usepackage[T1]{fontenc}
\usepackage{lmodern}
\def\BibTeX{{\rm B\kern-.05em{\sc i\kern-.025em b}\kern-.08em
    T\kern-.1667em\lower.7ex\hbox{E}\kern-.125emX}}

\newcommand{\Review}[1]{\textcolor{black}{#1}}
\newcommand{\Ishii}[1]{\textcolor{black}{#1}}

\newcommand{\Impl}[1]{\textsc{#1}}

\AtBeginDocument{
  \clearpage
  \onecolumn 
  \thispagestyle{empty}
  \vspace*{0pt} 

  \noindent
  {\LARGE IEEE Copyright Notice}

  \vspace{2em}

  \noindent
  Copyright (c) 2025 IEEE.

  \bigskip

  \noindent
  Personal use of this material is permitted. Permission from IEEE must be obtained for all other uses, in any current or future media, including reprinting/republishing this material for advertising or promotional purposes, creating new collective works, for resale or redistribution to servers or lists, or reuse of any copyrighted component of this work in other works.

  \vspace{3em}

  \noindent
  Accepted to be published in the 36th IEEE Intelligent Vehicles Symposium (IV), 2025.

  \vspace*{\fill}
  \twocolumn 
  \newpage
}

\begin{document}






\title{\LARGE \bf
Comparison of Lightweight Methods for Vehicle Dynamics-Based Driver Drowsiness Detection
}

\author{Yutaro Nakagama$^{1}$, Daisuke Ishii$^{1}$, and Kazuki Yoshizoe$^{2}$%
\thanks{\Ishii{This work was supported by JST, CREST Grant Number JPMJCR23M1, and JSPS, KAKENHI Grant Number 23K11969.}}
\thanks{$^{1}$Yutaro Nakagama and Daisuke Ishii are with School of Information Science, Japan Advanced Institute of Science and Technology, Ishikawa, Japan. 
        {\tt\small \{ynakagama, dsksh\}@jaist.ac.jp}}%
\thanks{$^{2}$Kazuki Yoshizoe is with Research Institute for Information Technology, Kyushu University, Fukuoka, Japan.
        {\tt\small yoshizoe@cc.kyushu-u.ac.jp}}%
}

\maketitle
\begin{abstract}
Driver drowsiness detection (DDD) prevents road accidents caused by driver fatigue.
Vehicle dynamics-based DDD has been proposed as a method that is both economical and high performance.
However, there are concerns about the reliability of performance \Review{metrics} and the reproducibility of many of the existing methods.
\Review{For instance, 
some previous studies seem to have a data leakage issue among training and test datasets, and 
many do not openly provide the datasets they used.}
%
\Review{To this end, this paper aims to compare the performance of representative vehicle dynamics-based DDD methods under a transparent and fair framework that uses a public dataset.}
We first develop a framework for extracting features from an open dataset \Review{by Aygun et al.} and performing DDD with lightweight ML models;
the framework is carefully designed to support a variety of configurations.
Second, we implement three existing representative methods and a concise \Ishii{random forest (RF)-based} method in the framework.
Finally, we report the results of experiments to verify the reproducibility and clarify the performance of DDD based on common metrics.
\Review{Among the evaluated methods, the RF-based method achieved the highest accuracy of 88\,\%.}
Our findings imply the issues inherent in DDD methods developed in a non-standard manner, and demonstrate a high performance method implemented appropriately.
\end{abstract}


\section{Introduction}

\emph{Driver drowsiness detection (DDD)} methods (Sect.~\ref{sec:method}) are essential to improve road safety by mitigating accidents caused by driver fatigue~\cite{Milter1988}.
In DDD, whether the driver is {drowsy} or {awake} is determined by the signals collected from a running vehicle.
As summarized in the surveys~\cite{Albadawi2022,Perkins2023}, the signals to be used are broadly classified into:
i)~\emph{vehicle dynamics} signals such as steering angle, vehicle speed, and lane offset;
ii)~\emph{behavioral} signals that are visual cues of the driver such as eye movements and facial expressions; and
iii)~\emph{biological} signals such as EEG (electroencephalogram) and heart rate.
Vehicle dynamics-based methods are advantageous for their ability 
to leverage existing vehicle sensors, enabling cost-effective, non-intrusive, and real-time DDD.

The state-of-the-art DDD methods are realized using (supervised) \emph{machine learning (ML)} models.
Compared to other domains, lightweight ML methods such as SVM (support vector machine) and RF (random forest) are often favored due to hardware limitations.
As good performance has been reported by existing methods, e.g, the accuracy of $98.12\,\%$~\cite{Arefnezhad2019} and $91.22\,\%$~\cite{Wang2022} and the correct rate of $95\,\%$~\cite{Zhao2009}, we focus on DDD using lightweight ML models in this paper.

Despite the promising results of the existing vehicle dynamics-based methods,
there appears to be a discrepancy between the reported results and the DDD functions implemented in real vehicles;
\Review{for instance, the vehicle dynamics signals described in \cite{Zhao2009} appear to be inconsistent with real-world vehicle behavior.}
We also noticed some inconsistencies between the reported performance and the results of our \Ishii{reproduction experiment (Sect.~\ref{sec:exp1})};
\Review{for example, while a $91.22\,\%$ accuracy was reported~\cite{Wang2022}, we were only able to confirm $76\,\%$; also, while an AUC of $0.97$ was reported~\cite{Arefnezhad2019}, it was necessary to use test data to achieve a similar level of performance.}
\Ishii{To this end, we suspect that there may be flaws such as data leakage~\cite{Kapoor2023} that undermine the reliability of reported results.}
One of the reasons could be a non-standard development process that does not follow the standard ML practice.
In many papers, the measured metric and the experimental configuration are not clearly described, so it becomes doubtful whether they are standard.
\Ishii{Another issue is the reproducibility of the methods;
most DDD implementations are not available due to their proprietary nature.}

\begin{table*}[t]
    \centering
    \caption{\Review{Overview of existing vehicle dynamics-based DDD methods.}}
    \label{tab_all_ddd}
    \begin{tabular}{|m{2.8cm}|m{3.0cm}|>{\centering\arraybackslash}m{4.7cm}|>{\centering\arraybackslash}m{2.0cm}|>{\centering\arraybackslash}m{3cm}|}
        \hline
        \centering \textbf{Citation} 
        & \centering \textbf{Metric} & \centering \textbf{Input Signal} & \!\!\!\!\!\! \centering \textbf{Ground Truth} \!\!\!\!\!\! & \textbf{ML Method} \\
        \hline\hline
        Arefnezhad et al.
        \cite{Arefnezhad2019} (\Impl{SvmA}) & 98\,\% (accuracy) & SWA, SWR & KSS & {SVM} \\
        \hline
        Arefnezhad et al.
        \cite{Arefnezhad2020} & 96\,\% (accuracy) & SWA, SWR, yaw rate, lateral accel., lateral deviation from centerline 
        & 
        Manual evaluation
        & {CNN, CNN-GRU, CNN-LSTM} \\
        \hline
        Zhao et al.
        \cite{Zhao2009} (\Impl{SvmW}) & 95\,\% (correct rate) & SWA 
        & 
        Event occurrence
        & {SVM} \\
        \hline
        Xue et al.
        \cite{Xue2023} & 94\,\% (accuracy) & SWA, SWR, lateral offset , lateral\&longitude speed, lateral\&longitude accel., throttle angle, headway distance & $n$-back task 
        & {SVM, RF} \\
        \hline
        Wang et al.
        \cite{Wang2022} (\Impl{Lstm}) & 91\,\% (accuracy) & SWR, lateral/longitudinal accel., speed, lane offset 
        & 
        Event occurrence
        & {Bi-LSTM, LSTM, SVM, $k$-NN, AdaBoost} 
        \\ 
        \hline
        Li et al.
        \cite{Li2017_2} & 88\,\% (accuracy) & SWA, yaw rate 
        & 
        Manual evaluation
        & {BPNN} \\
        \hline
        Krajewski et al.
        \cite{Krajewski2009} & 86\,\% (recognition rate) & SWA & KSS & {SVM, $k$-NN, DT, LR} \\
        \hline
        \!\!\! Eskandarian et al.
        \cite{Eskandarian2007} \!\!\! & 86\,\% (accuracy) 
        & SWA, SWR, steering wheel phase plot & SDR, PERCLOS & {ANN} \\
        \hline
        Mcdonald et al.
        \cite{Mcdonald2014} & 79\,\% (accuracy) & SWA & ORD & {RF, DT, NN, $k$-NN, NB, SVM, BT}\\
        \hline
        Mcdonald et al.
        \cite{Mcdonald2012} & 79\,\% (accuracy) & SWA & ORD & {RF} \\
        \hline
        Li et al.
        \cite{Li2017} & 78\,\% (accuracy) & SWA 
        & 
        Manual evaluation
        & {DTW} \\
        \hline
        Sandberg et al.
        \cite{Sandberg2008} & 
        75\,\% (average of sencitivity and specificity)
        & Speed, lateral position, SWA, yaw rate & KSS & {FFNN} \\
        \hline
        Chai et al.
        \cite{Chai2019} & 75\,\% (accuracy) & SWA & KSS & {MOL, SVM, BPNN} \\
        \hline
        Jin et al.
        \cite{Jin2012} & 74\,\% (accuracy) & Speed, accel., SWA, throttle position, yaw angle 
        & 
        \!\!\! Event occurence \!\!\!
        & {SVM} \\
        \hline
    \end{tabular}
    \begin{flushleft}
        SWA: Steering Wheel Angle, 
        SWR: Steering Wheel Rate, 
        KSS: Karolinska Sleepiness Scale, 
        SVM: Support Vector Machine, 
        CNN: Convolutional Neural Network, 
        GRU: Gated Recurrent Unit, 
        LSTM: Long Short-Term Memory, 
        RF: Random Forest, 
        $k$-NN: $k$-Nearest Neighbors, 
        DT: Decision Tree, 
        LR: Logistic Regression, 
        BPNN: Back Propagation Neural Network, 
        FFNN: Feedforward Neural Network, 
        MOL: Multiple Output Learning, 
        SDR: Sleep Deprivation Recognition, 
        PERCLOS: Percentage of Eyelid Closure, 
        ORD: Observer Rating of Drowsiness, 
        ANN: Artificial Neural Network, 
        DTW: Dynamic Time Warping, 
        NB: Naïve Bayes, 
        BT: Boosted Trees, 
        Bi-LSTM: Bidirectional Long Short-Term Memory, 
        AdaBoost: Adaptive Boosting, 
        BN: Bayesian Network.
    \end{flushleft} 
\end{table*}

This paper addresses vehicle dynamics-based DDD methods based on lightweight ML models.
\Ishii{Our objective is to compare the performance of representative methods under the same conditions using a publicly available dataset.}
Our contributions are twofold:
\begin{enumerate}
\item \emph{A common framework for implementing DDD methods} (Sect.~\ref{sec:method}).
    We design a DDD framework that is parameterized by
    ML pipeline configurations,
    sets of features extracted from input signals, 
    feature selection methods, and
    ML methods.
    \Ishii{The framework assumes the use of an open dataset (Sect.~\ref{sec:dataset}).}
    Based on the framework, we have implemented three existing methods~\cite{Arefnezhad2019,Zhao2009,Wang2022} and a simple RF method.
\item \emph{Experimental results}.
    The first experiment was conducted to verify the reproducibility of the existing DDD methods and revealed several issues in their development (Sect.~\ref{sec:exp1}).
    The second experiment compared the performance of our method implementations under a common configuration (Sect.~\ref{sec:exp2}).
    As a result, we identified a high-performance DDD method using reliable metrics.
\end{enumerate}

\section{Related Work}
\label{sec:related_work}

There are numerous papers on vehicle dynamics-based DDD~\cite{Albadawi2022,Perkins2023}.
Tab.~\ref{tab_all_ddd} summarizes a part of the existing work.
Each column shows the following information:
\Ishii{The citation};
the performance metrics reported in the papers;
input signals representing the vehicle dynamics;
the source of ground-truth labels, which are used to train ML models; and
ML methods to develop DDD models.
In this paper, we select three methods, i.e., \cite{Arefnezhad2019,Zhao2009,Wang2022}, with relatively good performance from among these and use them as the subject of our experiment.
We only select methods that use lightweight ML models and whose ground-truth labels are contained in the dataset we will use;
for example, we do not select \cite{Arefnezhad2020} because it uses a large-scale model, and \cite{Xue2023} because labeling is difficult with the dataset.
The three methods~\cite{Arefnezhad2019,Zhao2009,Wang2022} will be analyzed in more detail in Sect.~\ref{sec:four_methods}.


Several vehicle dynamics datasets have been developed.
\Review{These include PPB-Emo by Li et al. \cite{Li2022}, a multimodal dataset for emotion recognition in driving tasks; 
the dataset from the WACHSens project by Eichberger and Arefnezhad \cite{Eichberger2022}; 
and a dataset by Aygun et al.~\cite{Aygun2024}, 
the only open dataset that includes sufficient vehicle dynamics data and EEG data, and matches our use(Sect.~\ref{sec:dataset}).}
Other datasets have deficiencies in terms of vehicle data type, EEG data, event occurence data, recording frequency, open accessibility, etc.
%


{The models in \cite{Arefnezhad2019,Krajewski2009,Sandberg2008,Chai2019}} are trained using the KSS (Karolinska sleepiness scale), which is a subjective evaluation metric, as the source of the ground truth. 
In contrast, we employ EEG signals, objective physiological data capable of real-time monitoring of the driver's arousal state. 
DDD using EEG analyzes brain activity to assess alertness in real-time~\cite{Jung1997}. 
Since EEG provides a direct physiological measure of fatigue, it can detect drowsiness earlier than behavioral signs. 
ML models process EEG spectral features to classify alertness levels and trigger warnings, offering an objective approach to fatigue detection~\cite{Jap2009}.
%
However, implementing an EEG-based DDD 
requires intrusive and specialized equipment; 
in addition, it is prone to external vibrations and noise caused by unrelated behaviour.
Therefore, we only used the EEG data obtained in the experimental environment as the source of the ground truth.

%

%

\section{DDD Based on the Vehicle Dynamics Data}
\label{sec:method}


We formulate \emph{DDD} as a binary classification problem that classifies the driver's state as either \emph{drowsy} or \emph{awake}, based on vehicle dynamics data,
which are time series data sampled from the signals of \emph{steering wheel angle (SWA)} $\theta\,\mathrm{(rad)}$, \emph{steering wheel rate (SWR)} $\dot{\theta}\,\mathrm{(rad/s)}$, \emph{lateral velocity} $v_x\,\mathrm{(m/s)}$, \emph{longitudinal} and \emph{lateral accelerations} $a_x, a_y\,\mathrm{(m/s^2)}$, and \emph{lane offset} $\delta\,\mathrm{(m)}$. 
%
\begin{wrapfigure}[10]{r}{0.28\textwidth}
\centering
    \vspace{-1em}
    \hspace*{-2.5em}
    \includegraphics[width=0.35\textwidth,trim=10 70 10 80,clip]{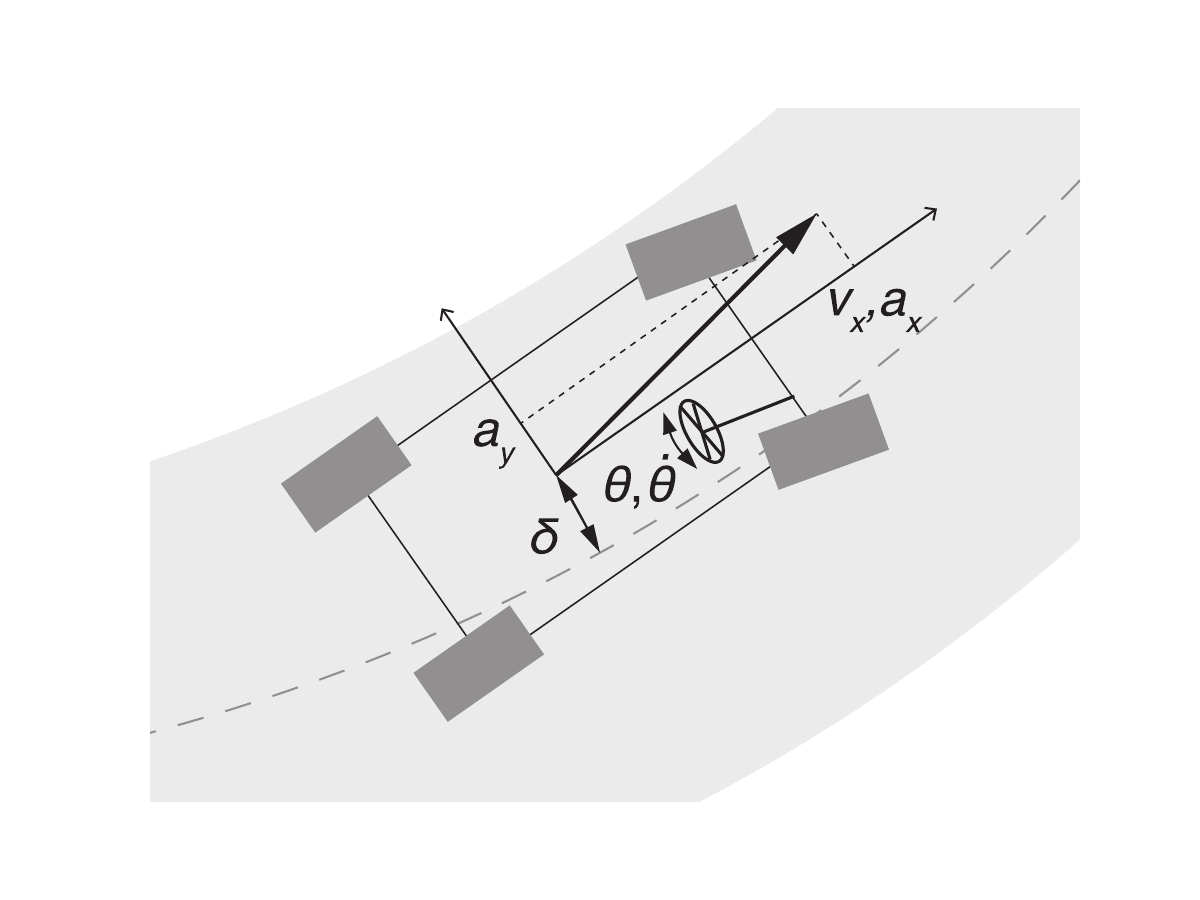}
    \vspace{-1em}
    \caption{The vehicle model.}
    \label{fig_veh_dyn}
\end{wrapfigure}
Fig.~\ref{fig_veh_dyn} illustrates the underlying vehicle model, which is basically a black box that outputs the signal data.
When using DDD, the data are processed as streams, but in this paper, we focus on the process of classifying the sample data from a time window independently from other time windows;
the process is parameterized by \emph{window size} and \emph{sampling frequency}.

\subsection{DDD Framework}

\Ishii{We propose a framework that extracts the common parts from lightweight DDD methods that use relatively small models based on basic ML techniques.}
An existing open dataset (Sect.~\ref{sec:dataset}) is used in common for training and evaluating the models.
We implement three methods~\cite{Arefnezhad2019,Zhao2009,Wang2022} that have  been reported to have good performance, and a basic classifier model for comparison.

\begin{figure}[t]
    \begin{minipage}[b]{.19\textheight}
   \centering
    \includegraphics[width=0.8\textwidth,trim=100 50 100 50,clip]{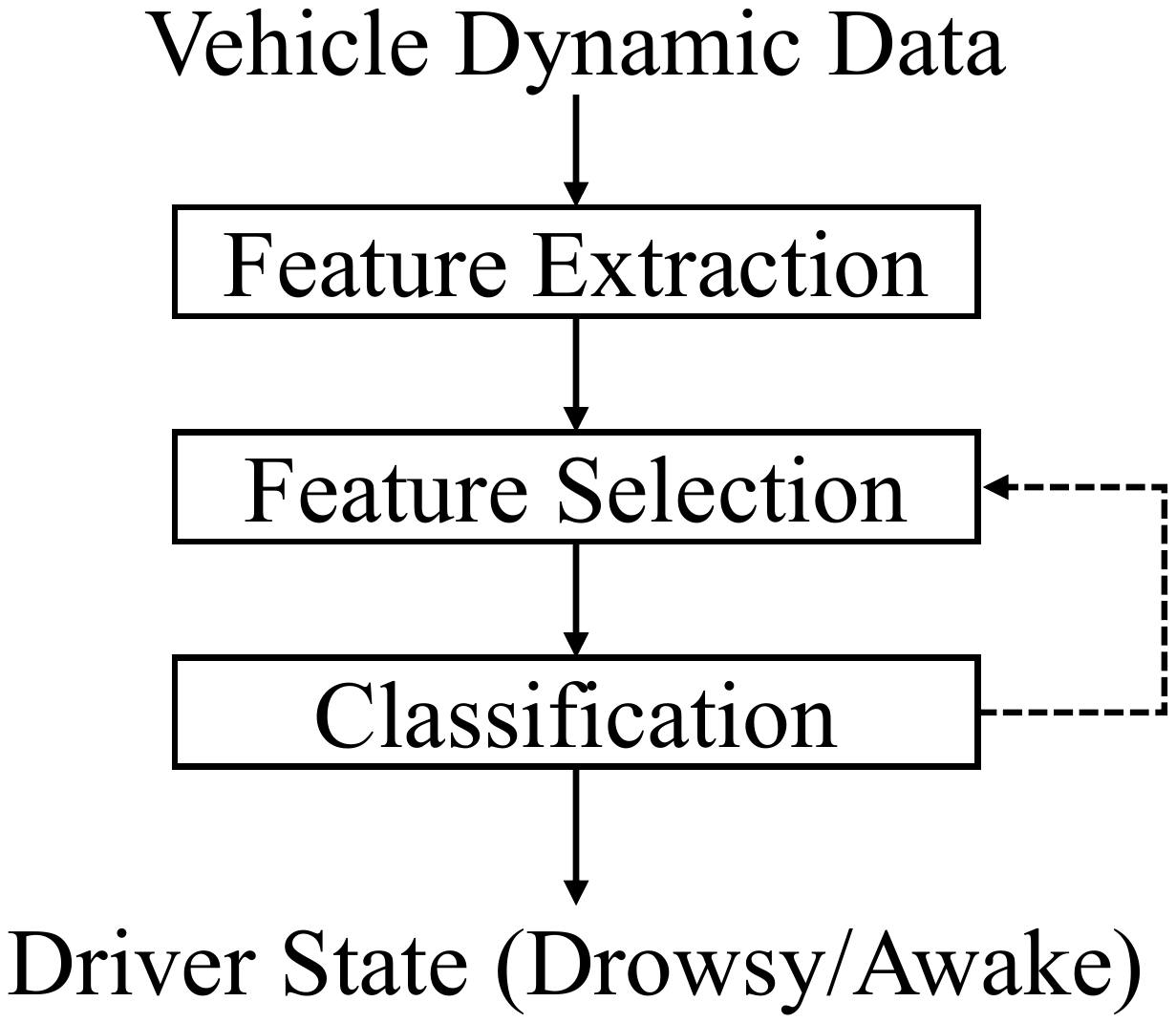}
    \caption{DDD process.}
    \label{fig:flowchart}
    \end{minipage}
    \begin{minipage}[b]{.15\textheight}
    \centering
    \includegraphics[width=0.9\textwidth,trim=100 0 100 0, clip]{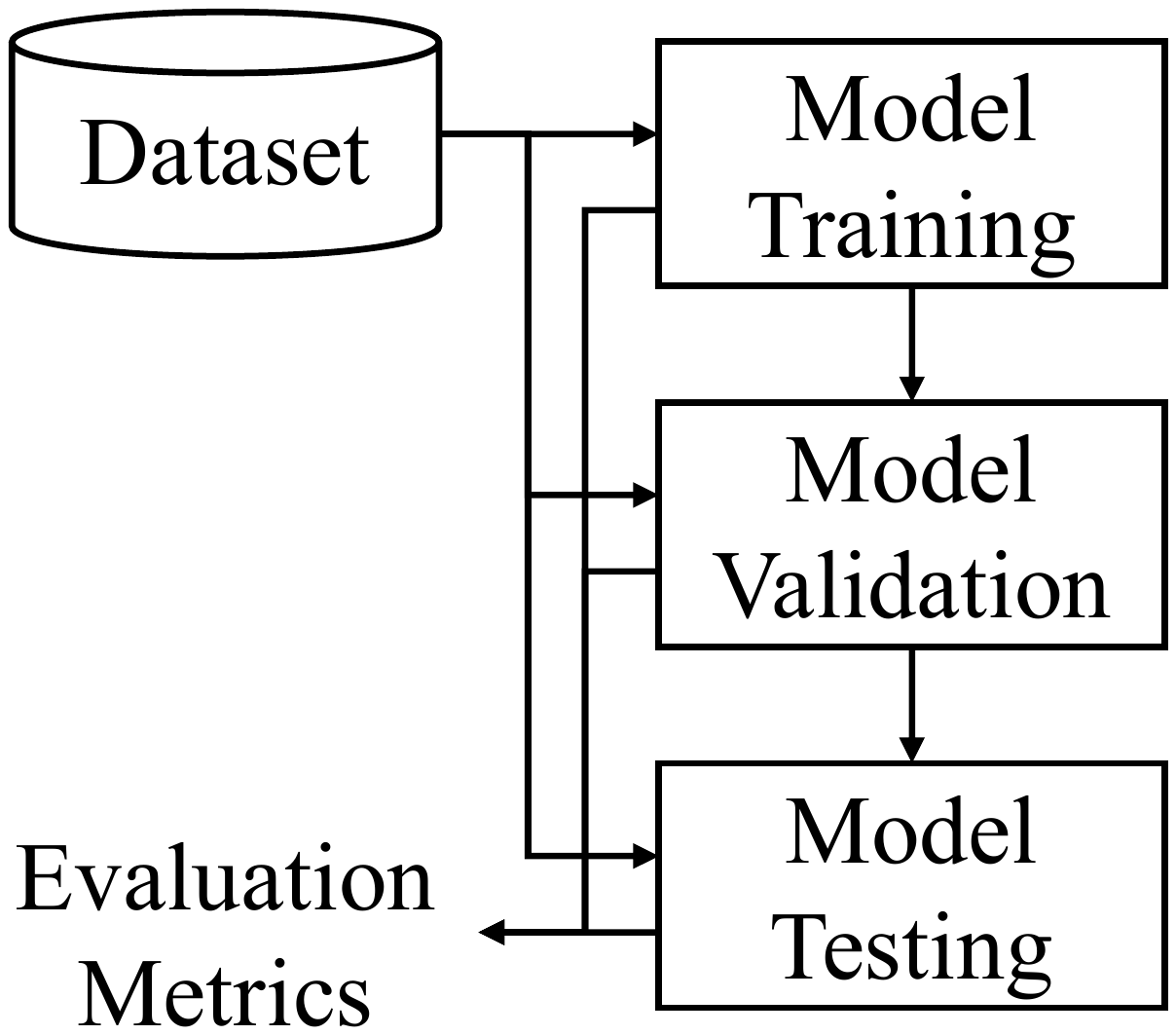}
    \caption{ML pipeline.}
    \label{fig:pipeline}
    \end{minipage}
\end{figure}

\Ishii{The common process for the DDD methods illustrated in Fig.~\ref{fig:flowchart} is executed on the framework; it consists of the following three steps.}
\begin{enumerate}
    \item \emph{Feature Extraction}. 
    The framework computes feature values from data sampled from a time window. These include statistical values (e.g. variance and fluctuation width) and values based on wavelets. We have prepared all of the features used by the three existing methods~\cite{Zhao2009, Arefnezhad2019, Wang2022}.
    A subset of features is predefined and used for each method.
    \item \emph{Feature Selection}. 
    The framework dynamically selects the likely features from a given set.
    It supports either of a standard selection method based on a statistical test or the methods used in \cite{Arefnezhad2019, Wang2022}.
    Some methods use a feedback DDD results.
    \item \emph{Classification}. The resulting set of features is input to a model to obtain the driver's state.
\end{enumerate}

\subsection{ML Pipeline Configurations}
\label{sec:conf}

The development of ML models follows the \emph{pipeline} depicted in Fig.~\ref{fig:pipeline}.
The dataset is usually split for each step, i.e., training, validation and testing; however, under a \emph{data leakage configuration}, a same subset may be used among steps in a development process.
In our framework, we characterize various \emph{configurations} of the ML pipeline by the following three elements:
\begin{itemize}
\item \emph{Data processing method} that specifies how the dataset is split for each pipeline step and which portion to use for the evaluation;
\item \emph{Labeling} that specifies the source of the ground-truth labels (i.e. driver states) and the granularity of the data (e.g. signal fragments for a time interval) to be labeled; and
\item \emph{Hyperparameters} that specify the window size, overlap ratio, and sampling rate.
\end{itemize}


\subsection{The MMDAP Dataset} \label{sec:dataset}

To implement the DDD methods, we use the \emph{multi-modal data acquisition platform (MMDAP) dataset} developed by Aygun et al.~\cite{Aygun2024,Scheutz2024}.
The dataset consists of time-series data, which record vehicle dynamics signals, road data, EEG signals and task event data required by our method;
they were collected in experiments using a driving simulator, dedicated sensors and measuring instruments.
The dataset is summarized in Tab.~\ref{tab:dataset}.

%
In an experiment session described in \cite{Scheutz2024}, a participant was instructed to drive on a straight and four-lane highway.
During the session, the participants were required to maintain an appropriate speed while adhering to the $65~\mathrm{mph}$ speed limit in the right lane.
In addition braking events and audio-based question events were systematically occured during the session.
The driving session included scenarios with and without the \emph{detection response task} (DRT), which required participants to press a button in response to tactile stimuli delivered every 6–10 seconds via a motor on their shoulder.
Each driving scenario covered a total distance of $37.4~\mathrm{km}$ and took approximately 25 minutes to complete. 
%

\begin{table}[t]
\centering
\caption{\Review{Specification of the MMDAP dataset~\cite{Aygun2024}.}}
\label{tab:dataset}
\begin{tabular}{|c|m{.31\textwidth}|} 
    \hline
    \multicolumn{1}{|c|}{\textbf{Category}} & \multicolumn{1}{c|}{\textbf{Description}} \\
    \hline
    \hline
    {\# Participants}
    & 82 (46\,\% female, 54\,\% male, avg. age: 20). \\
    \hline
    {Equipment} 
    & Medium fidelity partial-cab driving simulator, Enobio~8 EEG sensor system, etc.
    \\
    \hline
    {Recorded Data}
    & $a_x$, $a_y$, $v_x$, $\theta$, $\dot{\theta}$, $\delta$, DRT events, 
    brake events, 
    lane information, vehicle heading, and position. 
    ($60~\mathrm{Hz}$)
    \\
    \hline
    {EEG Data} 
    & 8~channels ($500~\mathrm{Hz}$)
    \\
    \hline
\end{tabular}
\end{table}

    
In our DDD framework, we use EEG data as a source of the ground-truth labels.
{We compute frequency-specific features for the EEG signal data from multiple channels.
Specifically, we analyze power spectral density in the $\vartheta$ ($4$--$8\,\mathrm{Hz}$), $\alpha$ ($8$--$13\,\mathrm{Hz}$), and $\beta$ ($13$--$20~\mathrm{Hz}$) bands~\cite{Horne2004}.
The ratio $(\vartheta+\alpha)/\beta$ is then derived to quantify drowsiness levels, based on \cite{Jap2009}. 
Following the labeling method in \cite{Arefnezhad2019}, the lower $60\,\%$ and the upper $22.2\,\%$ of the level distribution are classified as awake and drowsy, respectively.
}
    
We also use DRT event occurence data for ground-truth labeling.
The binary value of either during the task or a few seconds before and after it is interpreted as drowsy and awake, respectively.

\subsection{DDD Methods}
\label{sec:four_methods}

\begin{table*}[t]
\centering
\caption{\Review{Details of the three DDD methods}}
\label{tab_3_ddds}
\begin{tabular}
    {|c||c|c|c|}
    \hline
    \textbf{Method}
    & {\Impl{SvmA} (re-implementation of \cite{Arefnezhad2019})} 
    & {\Impl{SvmW} (re-implementation of \cite{Zhao2009})} 
    & {\Impl{Lstm} (re-implementation of \cite{Wang2022})} \\
    \hline
    \textbf{Input}
    & $\theta$, $\dot{\theta}$ ($60\,\mathrm{Hz}$) 
    & $\theta$ ($25\,\mathrm{Hz}$) 
    & $\dot{\theta}$, $v_x$, $a_x$, $a_y$, $\delta$ ($10\,\mathrm{Hz}$) \\
    \hline
    {\textbf{GT Source (orig.)}}
    & KSS 
    & Self-reported fatigue scores
    & \multicolumn{1}{c|}{Event occurence} \\
    \hline
    \!\!\!\!\! {\textbf{GT Source (re-impl.)}} \!\!\!\!\!
    & EEG
    & EEG
    & \multicolumn{1}{c|}{Event occurence} \\
    \hline
    {\textbf{\# Features}}
    & 36 (sample data statistical values)
    & 8 (wavelet frequency band energy)
    & 15 (modified signal data)
    \\ 
    \hline
    {\textbf{Selection Method}}
    & ANFIS, etc.
    & Not reported 
    & \multicolumn{1}{c|}{Student's $t$-test} \\
    \hline
    {\textbf{ML Method}}
    & SVM 
    & SVM 
    & Bi-LSTM with attention, etc.
    \\
    \hline
\end{tabular}
\end{table*}

We consider four DDD methods, \Impl{SvmA}, \Impl{SvmW}, \Impl{Lstm}, and \Impl{RF}.
Based on the common DDD framework, each method is specified by an ML method and a set of features.
The first three methods aim to replicate the existing methods that have been reported to have relatively high performance; we basically use the same ML methods and feature sets as the original methods.
%
%
In addition, the \Impl{RF} method is prepared as an integrated method using all the feature sets.

The details of the existing methods are shown in Tab.~\ref{tab_3_ddds}.
Each row corresponds to the name of the method,
the input signal(s) to be processed (with the processing frequency indicated), 
the source of the ground-truth labels (used by the original method),
the source used by the method re-implemented in this paper,
the number of extracted features from the input signals, the feature selection method, and the ML method to generate a classification model.

\vspace{.5em}

\subsubsection{\Impl{SvmA} (\Ishii{a method using SVM and ANFIS}; re-implementation of \cite{Arefnezhad2019})}

Arefnezhad et al.~\cite{Arefnezhad2019} propose a DDD method based on the 36 statistical features computed from the values of $\theta$ and $\dot{\theta}$ in a time window.
Statistics on time (e.g. range, energy, quartiles, and Shannon entropy) and frequency (e.g. variability, spectral entropy, and average power spectral density) are used.
%
The method calculates the importance of each feature using ANFIS (adaptive neuro-fuzzy inference system) and then performs a wrapper-based feature selection, which takes into account the previous classification results.
ANFIS is based on PSO (particle swarm optimization).
It uses an SMV model for the classification of the driver's states.

In the original work, ground truth labels are obtained from the KSS values surveyed for each time segment.
As mentioned previously, the re-implemented method in this paper uses labels based on the EEG in a uniform manner.

\vspace{.5em}

\subsubsection{\Impl{SvmW} (\Ishii{a method using SVM and wavelets}; re-implementation of \cite{Zhao2009})} \label{sec:ddd_svmA}

The method proposed by Zhao et al.~\cite{Zhao2009} applies wavelet decomposition on the SWA data ($\theta$).
8~features are obtained by a three-layer multiwavelet packet decomposition.
Especially Geronimo-Hardin-Massopust wavelet shows the best performance in \cite{Zhao2009}.
Among several wavelets, it has been reported that GHM (the Geronimo-Hardin-Massopust wavelet) shows the best performance;
thus, we also use GHM in our re-implementation.
The classification model is based on SVM. 

The original implementation is based on the ground truth obtained for each entire set of experiment.
Our re-implemented method uses EEG-based labels obtained for each time window.

\vspace{.5em}

\subsubsection{\Impl{Lstm} (\Ishii{a method using LSTM}; re-implementation of \cite{Wang2022})}

Wang et al.~\cite{Wang2022} proposes an LSTM-based method for time-series DDD.
It is based on the 3~feature signals, extracted from each of 5~input signals, i.e., $\dot{\theta}$, $v_x$, $a_x$, $a_y$, and $\delta$; a total of 15~feature signals are used.
The feature signals are obtained by applying each of the statistical processes called
``mean,'' ``standard deviation,'' and ``predicted error.''
%
The method then performs feature selection using $t$-test.
The classification is done with a bidirectional LSTM with an attention mechanism.

The original experiment was based on the scenario of receiving a phone call while driving.
DDD is a classification of whether (drowsy) or not (awake) the driver is responding to the call.
Accordingly, the ground truth is based on the time periods during which the call occurs.
In our re-implementation, we utilize the results using DRT events in the MMDAP dataset.
As original, the ground truth is based on whether or not an event occurs.

\vspace{.5em}

\subsubsection{\Impl{RF} (\Ishii{a method using random forest})}

\Review{We propose a simple 
DDD method based on a classification model generated with RF~\cite{Breiman2001}.
This method assumes the 6 vehicle dynamics signals and uses the features extracted using all previously mentioned methods. 
%
RF performs a certain degree of feature selection through its tree-based structure. 
To explicitly identify features with strong statistical relevance,
we additionally apply a univariate feature selection process, in which features are ranked by relevance using ANOVA F-values, 
The top-ranked features are then used to train the RF classifier. 
%
The model is trained and evaluated using EEG-based ground-truth labels. 
%
}


\subsection{Implementation}\label{sec:impl}

We implemented the DDD framework and the four methods in \Impl{Python}~3.10.12.
SVM and RF models were implemented using \Impl{scikit-learn}~1.5.2. 
\Impl{SelectKBest} was used for the feature selection of \Impl{RF}.
The LSTM model of \Impl{Lstm} was implemented using \Impl{TensorFlow}~2.18.0 and \Impl{Keras}~3.6.0. 
PSO of \Impl{SvmA} was implemented with \Impl{pyswarm}~0.6.
Additionally, we performed a hyperparameter tuning using \Impl{Optuna}~4.1.0 for each model.
\Review{The implementation is publicly available at \url{https://github.com/YutaroNakagama/vehicle_based_DDD_comparison}.}


%

\section{Experiment to Check the Reproducibility}\label{sec:exp1}

\begin{table*}[t]
\centering
\begin{threeparttable}[b]
    \caption{\Review{Result of reproducibility verification.}}
    \label{t:reprod}
    \begin{tabular}{|l|r|r|r | r | r|r|r|r|r|}
        \hline
        \multirow{3}{*}{}
        & \multicolumn{3}{c}{\bf \Impl{SvmA}}
        & \multicolumn{1}{|c}{\bf \Impl{SvmW}}
        & \multicolumn{5}{|c|}{\bf \Impl{Lstm}} \\
        \cline{2-10}
        & \multirow{2}{4.75em}{\bf ``AUC''} 
        & \multirow{2}{4.75em}{\!\!\!\!\!\bf ``Accuracy'' ($\%$)\!\!\!\!\!} 
        & \multirow{2}{4.75em}{\bf ``\# Features''}
        & \multirow{2}{4.75em}{\!\!\!\!\!\bf ``Correct Rate'' ($\%$)\!\!\!\!\!}
        & \multirow{2}{4.75em}{\!\!\!\!\!\bf ``Accuracy'' ($\%$)\!\!\!\!\!}
        & \multirow{2}{4.75em}{\!\!\!\!\!\bf ``Precision'' ($\%$)\!\!\!\!\!} 
        & \multirow{2}{4.75em}{\bf ``Recall'' ($\%$)}
        & \multirow{2}{4.75em}{\!\!\!\!\!\bf ``F1 Score'' ($\%$)\!\!\!\!\!}
        & \multirow{2}{4.75em}{\bf ``AUC''}
        \\
        & & & & & & & & & \\
        \hline
        \hline
        RV & $0.97$ & $98.12$ & $5$ 
        & $95$ 
        & $91.22$ & $94.59$ & $91.43$ & $92.924$ & $0.9740$ \\
        \hline
        C1 & $0.97$ & $97\phantom{.12}$ & $8$ 
        & $57$ 
        & $76\phantom{.22}$ & $76\phantom{.59}$ & $100\phantom{.43}$ & $86\phantom{.924}$ & $0.50\phantom{40}$ \\
        \hline
        C2 & $0.53$ & $65\phantom{.12}$ & $8$ 
        & $51$ 
        & $70\phantom{.22}$ & $84\phantom{.59}$ & $5\phantom{.43}$ & $10\phantom{.924}$ & $0.52\phantom{40}$ \\
        \hline
    \end{tabular}
    \begin{tablenotes}
        \item RV: reported values.
        \item C1: result using a configuration similar to the original.
        \item C2: result using a common configuration.
    \end{tablenotes}
\end{threeparttable}
\end{table*}

We have conducted an experiment to answer the following research question:
\emph{Can the performance metrics reported in previous studies be reproduced using our implementation and the MMDAP dataset?}

We used our implementation to reproduce the results close to the \emph{reported values (RVs)} in the literature.
In the experiment, the performance metric values obtained often differed considerably from RVs, probably because the metric or the configuration differed for details not described in the original papers.
In addition, it was often suspected that the configuration was not standard in some cases.
To this end, we checked the validity of the metric and configuration (Sect.~\ref{sec:conf}) used in each experiment by making a comparison between the two configurations:
\emph{C1}, a configuration that is supposed to be close to the original;
\emph{C2}, a common configuration to all experiments.


In C2, we split the dataset at a $8:1:1$ ratio for training, validation, and testing.
Driver states were derived from the EEG data in each window and labeled.
{We used the window size $3~\mathrm{seconds}$, overlap ratio $50~\mathrm{\%}$, and sampling rate $60~\mathrm{Hz}$.}

The RVs in the three papers and the results of our experiment with C1 and C2 are summarized in Tab.~\ref{t:reprod}.
In the following, we descibe how C1 was specified for each method and discuss the results.

\subsection{Reproducibility Verification with \Impl{SvmA}}

\subsubsection{Configuration}

We interpreted the metrics measured in \cite{Arefnezhad2019} as standard; they are AUC, accuracy, and the number of features (``\# features'') selected by the wrapper method.
\Review{Here, the ``number of features'' refers to the number of features finally selected from the full set during model training, and these selected features were used as inputs during evaluation.}

Since \cite{Arefnezhad2019} does not describe how the data are split, and the number of samples in their confusion matrix is almost 85\% of samples acquired from experiment,  
we assumed that the data had not been divided and that the \emph{training accuracy} was reported; we did the same in C1.
Labels had been derived from KSS but it was based on EEG in C1.
{The total experiment duration was approximately 164 hours in this study though only 20 hours and 36 minutes were utilized in \cite{Arefnezhad2019}.}
{The sampling rate $60~\mathrm{Hz}$, window size $3~\mathrm{s}$, and overlap ratio $50~\mathrm{\%}$ were set.}

\subsubsection{Discussion}

The RVs and the values obtained with C1 were almost comparable. 
However, a significant discrepancy was observed between the results with C1 and C2.
This is due to that the results with C1 were training accuracy.
%
%
We believe the model trained under C1 was suffering from overfitting.
{This is because optimizing the objective function for training accuracy often leads to overfitting on the training dataset, resulting in a loss of generalization performance.}
As usual, the test accuracy with C2 became lower.

\subsection{Reproducibility Verification with \Impl{SvmW}}

\subsubsection{Configuration}

In \cite{Zhao2009}, the metric ``\emph{correct rate}'' was evaluated.
Based on the analysis of the number of samples and that of misclassified samples reported in \cite{Zhao2009}, we interpreted it as accuracy (the percentage of correct results), which takes both the training and the test datasets into consideration.

According to \cite{Zhao2009}, the data was divided with a training-to-testing ratio of $3:7$ in C1.
For labeling, we used EEG data, although \cite{Zhao2009} used the participants' self-assessments.
The data were not processed with a sliding window, but the data for around 800~seconds was processed as a whole.
{The sampling rate was set to $60~\mathrm{Hz}$.}

\subsubsection{Discussion}

With both C1 and C2, the results were lower than RV.
We consider this is due to the adequacy of the source of the ground truth and evaluation methodology.
The results were worse with C2 than with C1.
We consider that it is more standard to measure using our ratio $training:validation:test=8:1:1$ for data splitting and the precision metric by test dataset for a classification problem like DDD. 

%

\subsection{Reproducibility Verification with \Impl{Lstm}} 

\subsubsection{Configuration}

As for \Impl{SvmA}, the metrics in \cite{Wang2022} were considered to be standard.
Precision, recall, F1 score were additionally used.
%
{A 5-fold cross-validation method was applied in C1.}
Driver states were classified according to whether or not the DRT events were occurred or not.
{The sampling rate $10~\mathrm{Hz}$, window size $10~\mathrm{second}$ without overlapping were set.}

\begin{figure}[t]
        \centerline{\includegraphics[width=0.5\textwidth, trim=0 5 0 0, clip]{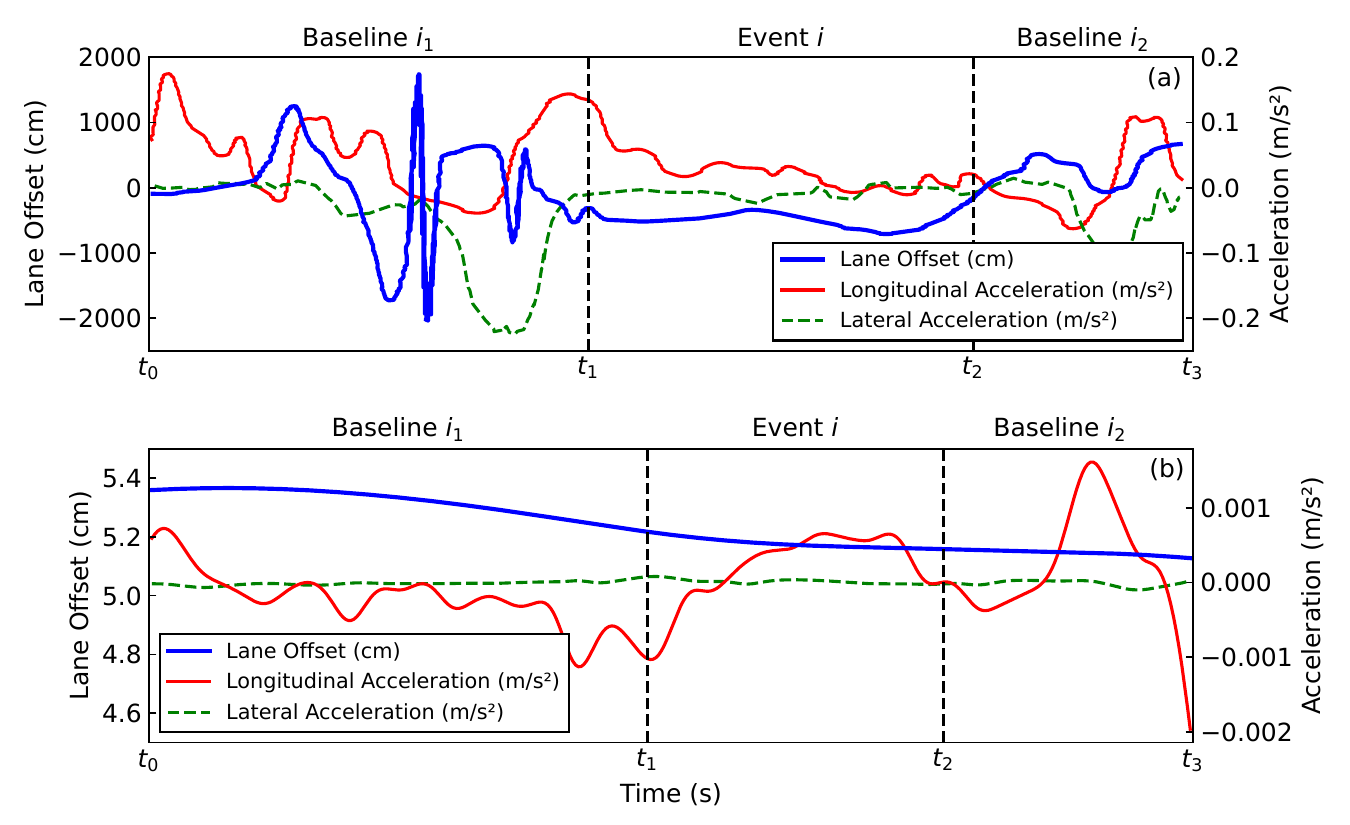}}
    \caption{Example input signals.}
    \label{wang_vs_ours}
\end{figure}

\subsubsection{Discussion}

The values we obtained were significantly lower than RVs.
Fig.~\ref{wang_vs_ours} shows both signal examples from \cite{Wang2022} and from the MMDAP dataset.
\Review{Fig.~\ref{wang_vs_ours}~(above) presents a sample from the dataset described in \cite{Wang2022}, where a clear change in signal patterns can be observed between the baseline and event intervals.} 
\Review{In contrast, Fig.~\ref{wang_vs_ours}~(below) shows an example from the MMDAP dataset, where such trend differences are not evident.} 
This may account for the gap between the results with C1 and the RVs.

{Achieving $100\,\%$ recall indicates that all positive instances are correctly identified; however, it may come at the cost of reduced precision. 
This could lead to a higher number of false positives. }
{In C1, event-based labeling was conducted following the original study, 
while C2 utilized EEG data to label driver alertness levels. 
The difference in labeling approaches revealed variations in model performance, even with the same model and dataset.}

\subsection{Summary}

In the experiment, we were only able to reproduce the reported performance partially under the standard interpretation of the metrics.
It was shown that some results reported in the literature probably reported the performance metrics of the training step, not the test step.
The result suggests that standardization of the evaluation process will be valuable.

\section{Evaluation of the \Impl{RF} Method}\label{sec:exp2}

The second experiment compared the four methods to answer the question:
\Ishii{\emph{How does the performance of the \Impl{RF} method compare to the other three methods?}}

\Review{While Sect.~\ref{sec:exp1} focused on reproducing and evaluating three existing methods, this section adds the proposed \Impl{RF} method into the comparison. By evaluating all four methods under the same configuration C2 and using standard performance metrics such as accuracy and AUC, we aim to provide a fair comparison among the methods and assess the effectiveness of the \Impl{RF} method.}

\subsection{Results}

\begin{figure}[t]
\centerline{\includegraphics[width=0.5\textwidth]{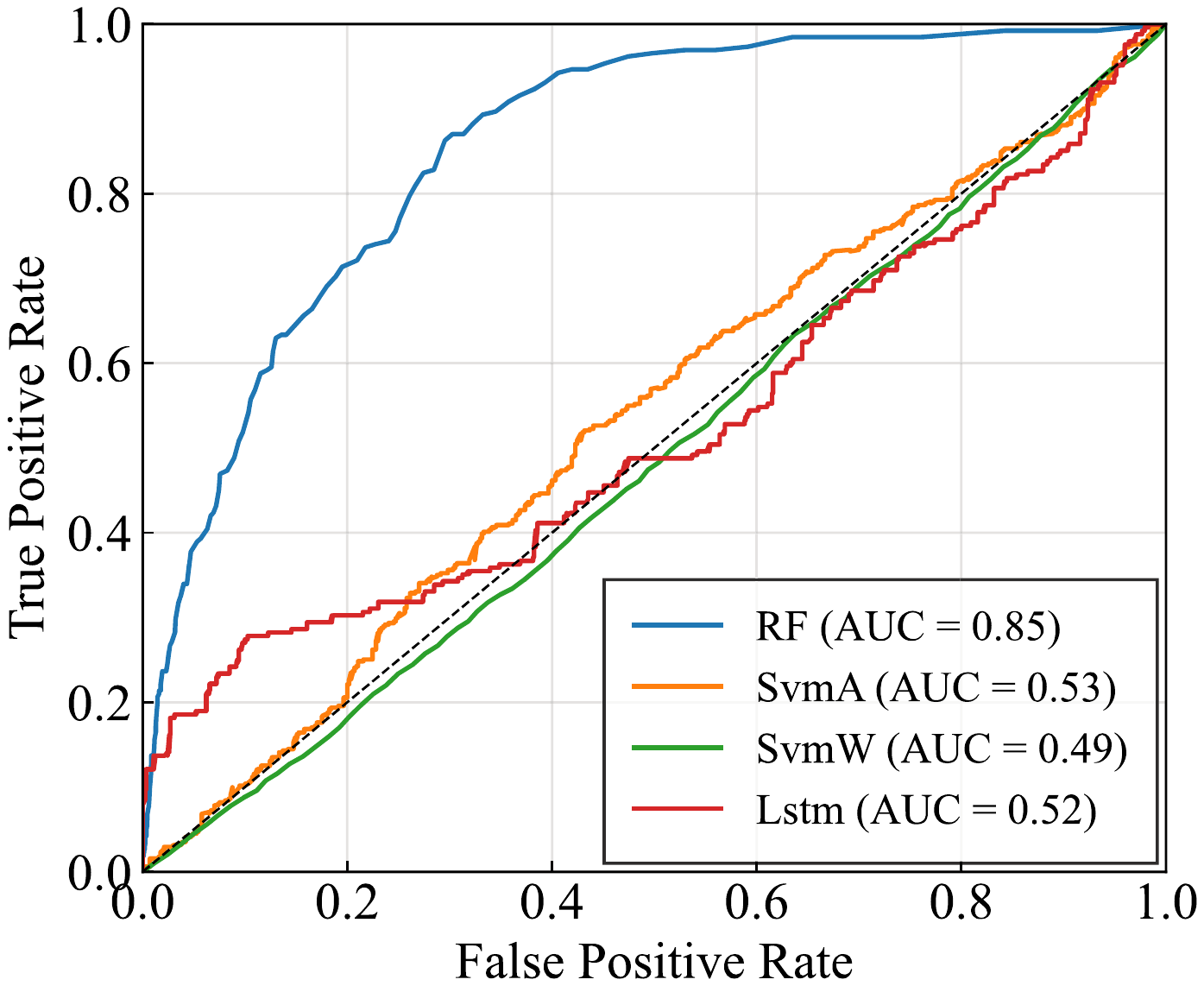}}
    \caption{ROC curves.}
    \label{roc}
\end{figure}

The results are shown in Tab.~\ref{tab:result_all} and Fig.~\ref{roc}.
Tab.~\ref{tab:result_all} shows the four metrics for each method.
Fig.~\ref{roc} illustrates the ROC curves.
\Review{As shown in Tab.~\ref{tab:result_all} and Fig.~\ref{roc}, \Impl{RF} achieved the best AUC (0.85) and accuracy (88\,\%) among the four methods. It also demonstrated a balanced recall (70\,\%) and reasonable precision (64\,\%), indicating both sensitivity and stability.}
Although \Impl{Lstm} achieved the best precision, the recall rate was relatively low.
\Review{In contrast, \Impl{Lstm} showed a recall as low as 5\%, which makes it less useful in practical applications.}

\begin{table}[t]
     \centering
     \caption{Performance comparison.}
     \label{tab:result_all}
     \begin{tabular}{|l|r|r|r|r|}
          \hline
          \multirow{2}{*}{\textbf{Method}} & \multirow{2}{*}{\textbf{AUC}} & \textbf{Accuracy} & \textbf{Precision} & \textbf{Recall} \\
          & & (\%) & (\%) & (\%) \\
          \hline \hline
          \Impl{RF} & $\mathbf{0.85}$ &  $\mathbf{88}$ & $64$ & $\mathbf{70}$ \\
          \hline
          \Impl{SvmA} & $0.53$ & $65$ & $65$ & $65$ \\
          \hline
          \Impl{SvmW} & $0.49$ & $49$ & $48$ & $49$ \\
          \hline
          \Impl{Lstm} & $0.52$ & $70$ & $\mathbf{84}$ & $5$ \\
          \hline
     \end{tabular}
\end{table}

\subsection{Discussion}

Overall, we consider that the performance of \Impl{RF} was the highest.
\Review{From a qualitative perspective, several factors can explain why \Impl{RF} outperformed the others.}
The primary reason for the high accuracy was that \Impl{RF} model utilized more features than other methods.
\Impl{SvmA}, trained using ANFIS, likely suffered from overfitting, which led to reduced test accuracy. 
In contrast, our proposed RF-based approach effectively avoided overfitting and gained better generalizability. 
As discussed in Sect.~\ref{sec:ddd_svmA}, \Impl{SvmW} used the entire dataset as input, unlike other methods that utilized time windows;
this approach may had led to lower accuracy because long time windows include irrelevant information, which makes it difficult to detect short-term variations. 
By leveraging time-windowed datasets, \Impl{RF} overcame this limitation and maintained higher performance.
\Impl{Lstm} may achieve high accuracy on datasets like those used by Wang et al., where signals exhibit significant changes before and after tasks. 
However, for datasets with minimal signal variation, such as the straight-driving dataset used in this study, the time window may have been too short for \Impl{Lstm} to effectively capture patterns.

\section{\Review{Threats to Validity}}\label{sec:threats}

\subsection{Hyperparameter Optimization}
The hyperparameters of RF, LSTM, and SVM may not have been equally optimized.
RF tends to perform well even with default settings, whereas LSTM and SVM require proper tuning.
If our implementation lacks sufficient optimization, \Impl{SvmA}, \Impl{SvmW} and \Impl{Lstm} may underperform, leading to an unfair evaluation.

\subsection{Feature Bias Towards RF}
The RF model utilizes all the features from other models, which may have given it an advantage.
If the other models used the same full set of features as RF, their performance could potentially improve.

\subsection{Dataset Generalizability}
The dataset used in this study consists of a driving task in a simulator on a straight road at a constant speed. 
While the road shape characteristics play a role, variations in task types could also influence model performance. Since the driving scenario involves minimal steering movements and little acceleration variation, the recorded vehicle dynamics signals may not fully capture real-world driving conditions. 
In more complex scenarios with curved roads or frequent acceleration and deceleration, these signals may behave very differently, potentially limiting the performance of \Impl{RF}.


\subsection{Individual Differences}
Driving behavior and the manifestation of drowsiness in EEG signals may vary among individuals.
Considering these individual differences among subjects may improve model accuracy.

\section{Conclusion}\label{sec:concl}

This study highlights the importance of reproducibility and standardized development process in the domain of DDD. 
As a result of a reproduction experiment using open data under a common framework, there were several cases where our results differed significantly from the results reported in the original papers.
\Review{In particular, the proposed RF-based method achieved the highest AUC of 0.85 and accuracy of 88\,\% among the four evaluated methods.}
As a bi-product, we have developed a concise RF-based DDD method that uses integrated feature set;
in comparative experiments using standardized performance metrics, it demonstrated the best performance.
Future issues include the development of DDD methods using larger-scale ML models, further evaluation experiments after preparing more diverse open datasets, and the deployment in actual vehicles to reduce fatigue-related accidents.



\end{document}